\documentclass[conference]{IEEEtran}
\usepackage[T1]{fontenc}
\IEEEoverridecommandlockouts

\usepackage{bm}
\usepackage{cite}
\usepackage{amsmath,amssymb,amsfonts}
\usepackage{algorithmic}
\usepackage{graphicx}
\usepackage{textcomp}
\usepackage{xcolor}
\usepackage{graphicx}
\usepackage{tikz}
\usepackage{pgfplots}
\usepackage{caption}
\usepackage{subcaption}
\usepackage{stfloats}

\def\BibTeX{{\rm B\kern-.05em{\sc i\kern-.025em b}\kern-.08em
    T\kern-.1667em\lower.7ex\hbox{E}\kern-.125emX}}
\begin{document}

\title{3D Extended Object Tracking based on Extruded B-Spline Side View Profiles}

\author{
\IEEEauthorblockN{
    Longfei Han\IEEEauthorrefmark{1} \IEEEauthorrefmark{2}, 
    Klaus Kefferp\"utz\IEEEauthorrefmark{1} \IEEEauthorrefmark{3}, 
    and J\"urgen Beyerer\IEEEauthorrefmark{2} \IEEEauthorrefmark{4}
}
\IEEEauthorblockA{\IEEEauthorrefmark{1} Application Center \guillemotright Connected Mobility and Infrastructure\guillemotleft , Fraunhofer IVI, Ingolstadt, Germany}
\IEEEauthorblockA{\IEEEauthorrefmark{2} Vision and Fusion Laboratory, Karlsruhe Institute of Technology (KIT), Karlsruhe, Germany}
\IEEEauthorblockA{\IEEEauthorrefmark{3} Technische Hochschule Ingolstadt, Ingolstadt, Germany}
\IEEEauthorblockA{\IEEEauthorrefmark{4} Fraunhofer IOSB, Karlsruhe, Germany}
}

\maketitle

\begin{abstract}
Object tracking is an essential task for autonomous systems. 
With the advancement of 3D sensors, these systems can better perceive their surroundings using effective 3D Extended Object Tracking (EOT) methods.
Based on the observation that common road users are symmetrical on the right and left sides in the traveling direction, we focus on the side view profile of the object.
In order to leverage of the development in 2D EOT and balance the number of parameters of a shape model in the tracking algorithms, we propose a method for 3D extended object tracking (EOT) by describing the side view profile of the object with B-spline curves and forming an extrusion to obtain a 3D extent.
The use of B-spline curves exploits their flexible representation power by allowing the control points to move freely. 
The algorithm is developed into an Extended Kalman Filter (EKF). 
For a through evaluation of this method, we use simulated traffic scenario of different vehicle models and real-world open dataset containing both radar and lidar data. 
\end{abstract}


\section{Introduction}
The ability to track the states of the objects around an autonomous system is crucial for extending its ability to interact with the surroundings. 
For autonomous driving \cite{ab3dmot} or intelligent transportation systems \cite{flexsense}, where perception is at the core of the functionality, object tracking lays the foundation for further tasks such as maneuver planning or traffic flow optimization. 
With the recent development of high resolution sensors like lidar or radar, the objects could be measured with more than one measurement, paving the way for the development of Extended Object Tracking (EOT), where the states of the objects could be described not only by a point along with its kinematic states, but also with extent information \cite{eot}. 

EOT has been an active research area for more than two decades. 
Many efficient trackers have been proposed. 
The shape models play a tremendous role in the tracker. 
At the early age, the shape model has been focused on 2D with possible extension in the 3rd dimension \cite{RandomMatrix, RHM, 3deot}.
These models are typically primitive geometries such as spheres, ellipsoids, or cubes \cite{factorgraph}. 
They only serve to represent the shape to a certain extent. 
To further represent the object shape with details, new models have been introduced \cite{GP, RHMStar}, which fall into the star convex shape category and could work flexibly with top view projections of objects for different road users.
However, when working with these models to extrude the top view of an object into a 3D shape, the model suffers from mismatch at positions such as the bonnet, leading to large artificial measurement error. 
On the other hand, the extension of these models in 3D needs a special treatment, since the representation of the objects evolves from polar coordinate system to spherical coordinate systems \cite{3dgp, SH, SDS, HCC, FCDS}. 
These could represent the shape with high accuracy, but at the expense of a much higher number of parameters. 
\begin{figure}[t]
    \centering
        \begin{tikzpicture}
        \node[anchor=south west,inner sep=0] (image) at (0,0) 
            {\includegraphics[width=\linewidth]{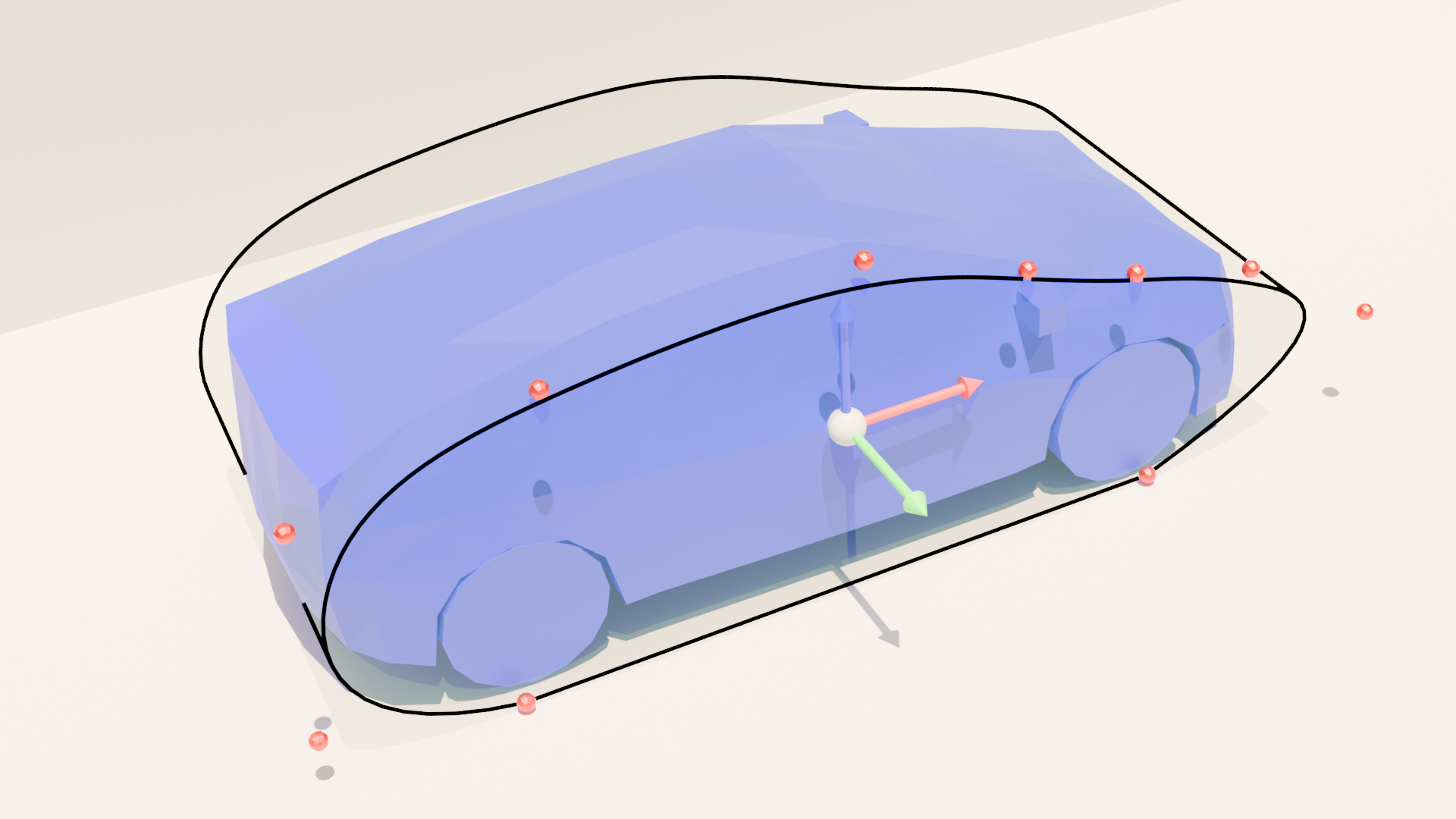}};
        \begin{scope}[x={(image.south east)},y={(image.north west)}]
            \draw[blue, thick, ->] (0.45, 0.1) node[black, below right=-2pt] {Control Points $\bm{x}_c$} -- (0.38,0.13);
            \draw[blue, thick, ->] (0.45, 0.4) node[black, below left=-2pt] {$\bm{x}_m$}-- (0.55,0.46);
            \draw[blue, thick] (0.2,0.73) -- (0.2,0.83);
            \draw[blue, thick] (0.3,0.43) -- (0.3,0.53);
            \draw[blue, thick] (0.2,0.78) -- (0.3,0.48) node[black, midway, right] {width $q$};
            \draw (0.75,0.55) node[black] {Caps};
            \draw (0.65,0.75) node[black] {Extrusion};
        \end{scope}
    \end{tikzpicture}
    \caption{The concept of extruding a side view profile into a 3D shape. 
    The black contour lines represent the estimated model. 
    The blue mesh is the true mesh of the object obtained in CARLA \cite{carla}. 
    The 3 axis represents the body frame of reference of the object. 
    The red points defined in the body frame are the control points of the B-spline curve. 
    They can be freely placed.}
    \label{fig:concept}
\end{figure}
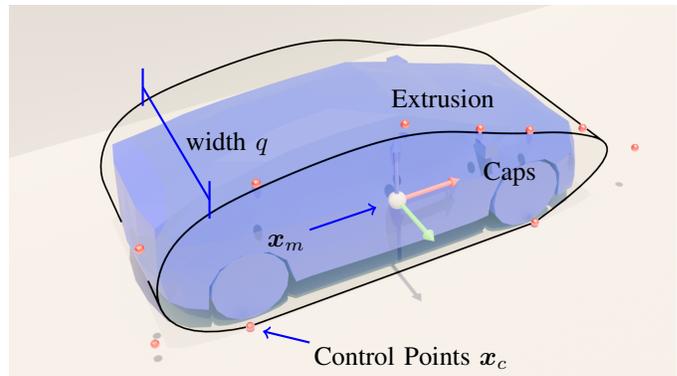

In order to cope with the development of 3D sensors, or even 4D sensors if the Doppler measurement is considered as an additive dimension as in automotive radars, while trying to find a balance in the increase of the number of parameters, we propose to track a 3D extended object as an extrusion of the side view profile.
For common objects on the road, the symmetry of the left and right side of the object in the direction of motion has been observed.
The side-view profile serves as a good representation of the objects. 
However, the profile is not necessarily star-convex, given the level of detail one chooses to model it.
Therefore, we use the B-spline curve as the shape model.
\cite{bspline} has shown that one can use B-spline curves to represent a box-shaped representation of a vehicle. 
Their work inspires the use of B-spline curves in EOT, but limits the flexibility of B-spline curves by setting semi-fixed control points of the spline and varying the scale only in x and y direction. 
This work has been improved by algorithms aimed at associating measurements and the walk parameter along the spline \cite{improvedbspline}.
Similar to this effort, \cite{NURBS} addresses the problem of using non-uniform rational B-splines (NURBS) surface to represent the objects in 3D. 
They show the power of using NURBS in EOT, but also limit the modeling capability by using the scales in x,y,z direction with or without the weights of the control points.
In contrast, we reduce the number of parameters in the width direction of the vehicle and focus on the side view. 
By doing so, we also allow the control points to vary freely, so that the shape could be captured with better accuracy. 
This also marks a difference to a pioneering work in the extrusion of a 2D profile into a 3D shape \cite{3dextrusion}, where the shape is defined by the extrusion of circles into a bottle, which mainly happens in the top view. 

To summarize our contribution in this paper:
\begin{enumerate}
    \item We propose a shape model for EOT to represent an object with a description of its side view profile as a B-spline curve,
    \item We define the process model and the measurement model and derive the tracker as an Extended Kalman Filter (EKF),
    \item We validate this method with different vehicles in both CARLA simulation and an open dataset, with both lidar and radar data. 
\end{enumerate}
The rest of the paper is organized as follows: Section \ref{Chapter2: Preliminary} gives background information about B-spline curves. The EOT problem is then defined in section \ref{Chapter3: ProblemFormulation}. The EKF tracking algorithm is described in Section \ref{Chapter4: Implementation}. In Section \ref{Chapter5: Evaluation} we show the through evaluation of the method. Finally, Section \ref{Chapter6: Conclusion} concludes the paper and gives further research directions.
\section{Preliminaries}
\label{Chapter2: Preliminary}
This section provides a brief introduction to B-spline curves.
Given a set of control points $\bm{\underline{c}}=(\bm{c}_i)_{i=1}^n$\footnote{Bold symbols denote vectors, while underlined symbols denote sets of points.} and a knot vector  $\bm{\tau}=(\tau_i)_{i=1}^{n+d+1}$, a B-spline curve $\bm{s}(\tau)$ of degree $d$ is defined as:
\begin{equation}
\label{eq: spline}
    \bm{s}(\tau) = \sum_{i = 1}^{n} B_{i,d}(\tau)\bm{c}_{i},
\end{equation}
where $\tau$ is a "walking parameter" along the spline curve.
The function $B_{i,d}(\tau)$ is given by
\begin{equation}
    B_{i,d}(\tau) = \frac{\tau-\tau_i}{\tau_{i+d}-\tau_{i}}B_{i,d-1}(\tau) + \frac{\tau_{i+1+d}-\tau}{\tau_{i+1+d}-\tau_{i+1}}B_{i+1,d-1}(\tau).\end{equation}
In cases of  ``divided by zero'' the function $B_{i,d}$ returns zero.
This basic function $B_{i,d}(\tau) = B_{i,d, \bm{\tau}}(\tau)$ is called a B-spline of degree $d$ (with knots $\bm{\tau}$).
This formulation benefits the problem description for the shape modeling, since the recursion is written in the B-splines, leading to an explicit representation of the control points $\bm{c}$.  
We refer the reader to \cite{splinemethod} for more properties and algorithms related to B-splines.
Furthermore, a more practical way to deal with the recursion is to expand the equation and write it in a matrix form \cite{splinemethod}.
In this case, we have $\bm{U}_p^\mu(\tau) = \bm{U}_p(\tau)$ in Eq. (\ref{eq: U}), where  $d+1 \leq \mu \leq n$, and $0<p \leq d$. 
\begin{figure*}[h]
    \centering
\begin{equation}
\label{eq: U}
	\bm{U}_p(\tau)=\begin{bmatrix}
		\frac{\tau_{\mu+1}-\tau}{\tau_{\mu+1}-\tau_{\mu+1-p}} & \frac{\tau-\tau_{\mu+1-p}}{\tau_{\mu+1}-\tau_{\mu+1-p}} & 0 &\cdots & 0 \\
		0 & \frac{\tau_{\mu+2}-\tau}{\tau_{\mu+2}-\tau_{\mu+2-p}} & \frac{\tau-\tau_{\mu+2-p}}{\tau_{\mu+2}-\tau_{\mu+2-p}} & \cdots & 0 \\
		\vdots & \vdots & \ddots & \ddots & \vdots \\
		0 & 0 & \cdots & \frac{\tau_{\mu+p}-\tau}{\tau_{\mu+p}-\tau_{\mu}} & \frac{\tau-\tau_{\mu}}{\tau_{\mu+p}-\tau_{\mu}} 
	\end{bmatrix}
\end{equation}
\end{figure*}
We call $\bm{U}_p(\tau)$ the B-spline matrix.
The B-splines $\{B_{j,d}\}^\mu_{j=\mu-d}$ (of degree $d$) in the interval $\tau \in [\tau_\mu, \tau_{\mu+1})$ can then be written as 
\begin{equation}
\label{eq: matrixmultiplication}
\begin{split}
	\bm{B}_d &= \begin{bmatrix}B_{\mu-d,d} & B_{\mu-d+1,d} & \ldots & B_{\mu,d}\end{bmatrix} \\
    &=\bm{U}_1(\tau)\bm{U}_2(\tau)\cdots\bm{U}_d(\tau).
\end{split}
\end{equation}
The spline curve in this interval can then be defined as 
\begin{equation}
\label{eq: recursive bspline}
        \bm{s}(\tau) = \bm{U}_1(\tau)\bm{U}_2(\tau)\cdots\bm{U}_d(\tau){\bm{c}}_d.
\end{equation}
Here ${\bm{c}}_d = [\bm{c}^\top_{\mu-d}, \bm{c}^\top_{\mu-d+1}, \cdots, \bm{c}^\top_\mu ]^\top$.

Now that we have introduced the B-spline curves, we treat the task of finding the side view profile as a curve fitting problem. 
Since the tracking follows the Bayesian filter, we are interested in a recursive solver for the curve fitting problem. 
There have been works such as \cite{kalmanspline, kalmanspline2} that deal with spline curve fitting with both fixed range and extending range. 
In this work, we add Eq (\ref{eq: recursive bspline}) to the shape representation and derive the EKF in a complete way.

\section{Problem Formulation}
\label{Chapter3: ProblemFormulation}
In this section, we define the tracking problem and derive the shape model within the EOT tracker. 
\subsection{State of the Extended Object}
We first define the state of the object $\bm{x}_k$  at the time step $k$ as
\begin{equation}
   \bm{x}_k= [\bm{x}_{k,m}^\top, \bm{x}_{k,e}^\top]^\top 
\end{equation}
The subscript time index $k$ will be neglected in the description of states for simplicity.
The kinematic state $\bm{x}_m$ is defined as 
\begin{equation}
    \bm{x}_m = [x_x, x_y, v_{xy}, \psi, \omega, x_z, v_z]^\top,
\end{equation}
where the position of the reference point of the object in 3D is represented with $[x_x, x_y, x_z]^\top$.
The speed of the object moving in the 2D ground plane is represented with $v_{xy}$, and the speed in the vertical direction is shown in $v_z$.
The orientation of the object plays an important role along with the angular velocity and is represented by $\psi$ and $\omega$. 
The kinematic state $\bm{x}_m$ defines the body frame of reference in the Fig. \ref{fig:concept}.
For the extent information, $\bm{x}_e$ is used.
\begin{equation}
 \bm{x}_e = [q, x_{c^1_x}, x_{c^1_z}, \cdots, x_{c^n_x}, x_{c^n_z} ]^\top,
\end{equation}
where $q$ is the width of the object. 
The variable $\underline{\bm{x}}_c =(\bm{x}_{c^i})_{i=1}^n=([x_{c^i_x}, x_{c^i_z}]^\top)_{i=1}^n $ represents the set of control points of the 2D B-spline curve in the $xz$ plane, which are represented as the red spheres in Fig. \ref{fig:concept}. 
They could be defined independently of $y$ in the body frame and are moved to the $y_{max}$ plane in the figure for demonstration purposes.
The elements in $\bm{x}_e$ form the black outline in the Fig. \ref{fig:concept}.

\subsection{Shape Model}
\label{chapter: shapemodel}
The shape of the object is defined as an extrusion of the side view profile described by a B-spline curve. 
It is described in two parts, the extrusion and the caps at the start and end of the extrusion. 
For the extrusion part, we describe the points on the B-spline curve, neglecting the length in the extrusion direction.
For the caps, we describe only the y element of the points, neglecting their position in the $xz$ plane.

We further define the shape model in its body frame of reference.
All possible points $\bm{z}^L$ on the surface of the shape are
\begin{equation}
\label{eq: shapemodel}
    \bm{z}^L(\bm{x}_e, \tau) =  \begin{cases}
        \left[z^L_x(\tau),z^L_y, z^L_z(\tau) \right]^\top,  \quad z^L_y \in \left[-\frac{q}{2}, \frac{q}{2}\right],\\
        \left[z_x^L, z_y^L, z_z^L\right]^\top, \quad z_y^L \in \left\{ \pm \frac{q}{2} \right\} .
    \end{cases}
\end{equation}
For the points on the extrusion, we have
\begin{equation}
\left[z^L_x(\tau),  z^L_z(\tau)\right]^\top= \bm{s}(\tau)
\end{equation}
Combining with Eq. (\ref{eq: spline}), we get
\begin{equation}
\label{eq: local measurement source}
        \left[z^L_x(\tau),  z^L_z(\tau)\right]^\top =  \sum_{i = 1}^{n} B_{i,d}(\tau)\bm{x}_{c^i}. 
\end{equation}
The non-zero terms in $(B_{i,d})_{i=1}^n$ are calculated with Eq. (\ref{eq: matrixmultiplication}).

For the points on the caps, we have \begin{equation}
\left[z^L_x,z^L_z\right] \in \Gamma = \left\{\left[z^L_x,z^L_z\right] \ | \ g(z^L_x,z^L_z) \leq 0\right\}    
\end{equation}
as the set of interior points on the cap, $g$ denotes a signed distance function. 

The width $q$ can be calculated from the difference between the points on the two caps. 
In this work, we set the width to known values in advance, since the width of vehicles does not cover a large range.

\section{Implementation}
\label{Chapter4: Implementation}
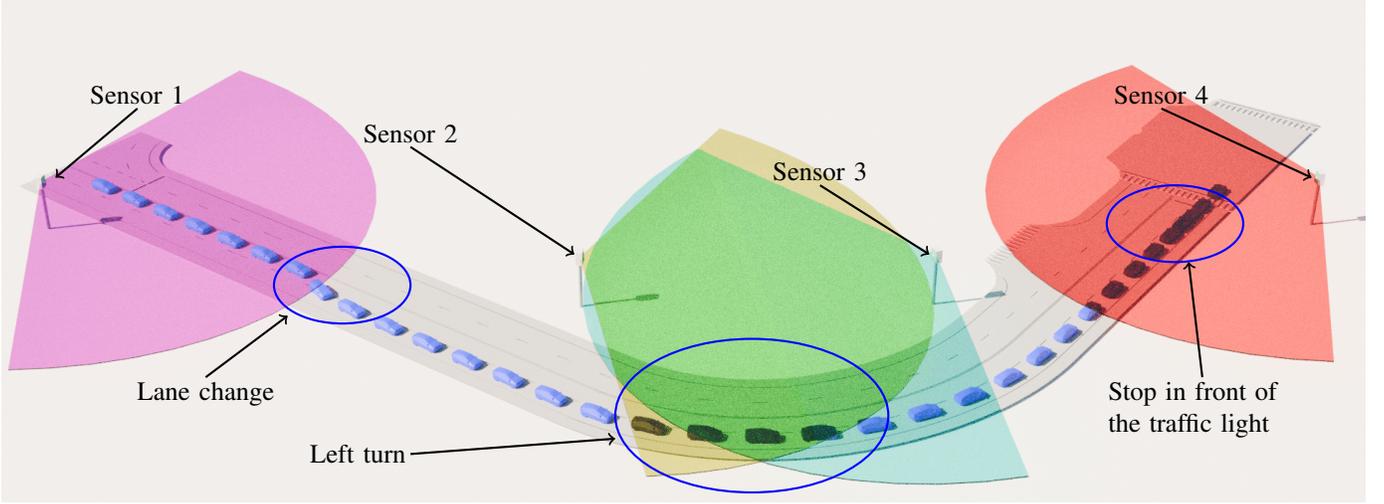
\begin{figure*}[h]
    \centering
\begin{tikzpicture}
            \clip (0,40pt) rectangle (\linewidth,230pt);
        \node[anchor=south west,inner sep=0] (image) at (0,0) 
            {\includegraphics[width=\linewidth]{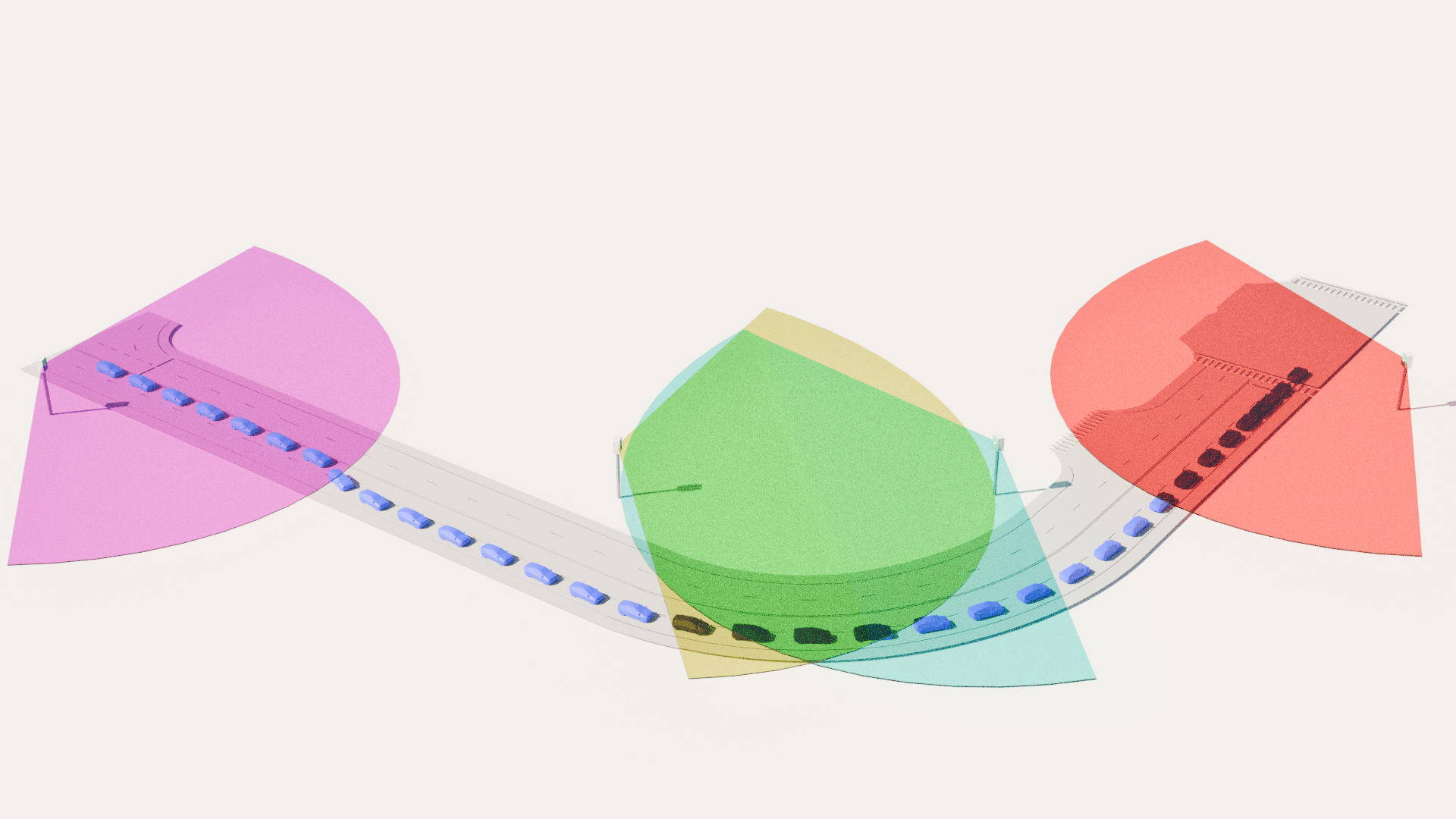}};
        \begin{scope}[x={(image.south east)},y={(image.north west)}]      
            \draw[black, thick, ->] (0.1, 0.65) node[above=-2pt] {Sensor 1} -- (0.04, 0.56);
            \draw[black, thick, ->] (0.3, 0.6) node[above=-2pt] {Sensor 2} -- (0.42, 0.46);
            \draw[black, thick, ->] (0.6, 0.55) node[above=-2pt] {Sensor 3} -- (0.68, 0.46);
            \draw[black, thick, ->] (0.85, 0.65) node[above=-2pt] {Sensor 4} -- (0.96, 0.56);
            \draw[black, thick, ->] (0.15, 0.3) node[below=-2pt] {Lane change} -- (0.21, 0.38);
            \draw[blue, thick] (0.25, 0.42) circle (0.05);
            \draw[black, thick, ->] (0.3, 0.2) node[left=-2pt] {Left turn} -- (0.45, 0.22);
            \draw[blue, thick] (0.55, 0.25) circle (0.1);
            \draw[black, thick, ->] (0.88, 0.3) node[below=-2pt, text width=2.5cm] {Stop in front of the traffic light} -- (0.87, 0.45);
            \draw[blue, thick] (0.86, 0.5) circle (0.05);
        \end{scope}
    \end{tikzpicture}
    \caption{The setup of the simulative environment for the evaluation.
    Four sets of sensors are placed at the position shown.
    Their field of view is represented by the colored area.
    A vehicle moves from the left to the right side.
    The driving behavior along the trajectory is marked in the figure.}
    \label{fig:setup}
\end{figure*}
In this section, we define the process model and the measurement model for deriving the EKF for tracking.
\subsection{Process Model}
The process model is a combination of a constant turn rate and velocity (CTRV) \cite{comparison} model for the motion in the 2D ground plane and a constant velocity (CV) model for the vertical motion. 
The planar motion uses two sets of equations based on the angular velocity of the object $\omega$. 
When $\omega$ is considerably large, we have
\begin{equation}
\begin{split}
    &\bm{x}_{k+1,m}  
    =\begin{bmatrix}
    x_{k,x} + \frac{v_{k,xy}}{\omega_k} (-\mathrm{s}({\psi_k}) +\mathrm{s}{(\psi_k+\Delta{t}\omega_k})) \\
    x_{k,y} + \frac{v_{k,xy}^t}{\omega_k} (\mathrm{c}({\psi_k}) - \mathrm{c}{(\psi_k+\Delta{t}\omega_k}))\\
    v_{k,xy} \\
    \psi_k + \Delta{t} \omega_k\\
    \omega_k \\
    x_{k,z} + \Delta{t} v_{k,z}\\
    v_{k,z}\\
    \end{bmatrix} + \bm{w}_m.
\end{split}
\end{equation}
The $\mathrm{s}(\cdot)$ and $\mathrm{c}(\cdot)$ indicate the $\sin$ and $\cos$ functions. 
When the $\omega$ is close to $0$, we have 
\begin{equation}
\begin{split}
    &\bm{x}_{k+1,m} 
    =\begin{bmatrix}
    x_{k,x} + \Delta{t} \cdot v_{k,xy}\cdot \mathrm{c}({\psi_k}) \\
    x_{k,y} + \Delta{t} \cdot v_{k,xy} \cdot \mathrm{s}({\psi_k}) \\
    v_{k,xy} \\
    \psi_k + \Delta{t} \omega_k\\
    \omega_k \\
    x_{k,z} + \Delta{t}v_{k,z}\\
    v_{k,z}\\
    \end{bmatrix} + \bm{w}_m.
\end{split}
\end{equation}
Here $\bm{w}_m \sim \mathcal{N}(\bm{0}, \bm{Q}_m)$ is the process noise for the kinematic state ($\bm{Q}_m = \mathrm{diag}(\sigma_x^2, \sigma_y^2, \sigma^2_{v_{xy}}, \sigma^2_\psi, \sigma^2_\omega, \sigma^2_z, \sigma^2_{v_z} )).$
For the extent information, we have 
\begin{equation}
    \bm{x}_{k+1,e} = I \  \bm{x}_{k,e} + \bm{w}_e.
\end{equation}
$\bm{w}_e \sim \mathcal{N}(\bm{0}, \bm{Q}_e)$ is the process noise for the extent state, with $\bm{Q}_e=\mathrm{diag}(\sigma_q^2, \sigma_{c_x^1}^2, \sigma_{c_z^1}^2, \cdots, \sigma_{c_x^n}^2, \sigma_{c_z^n}^2)$
A forgetting factor could be added. 

\subsection{Measurement Model}
Following Section \ref{chapter: shapemodel}, the measurement model is defined in the body frame of reference.
At time $t$, we get $m$ measurements $(\bm{y}_{k,l})^m_{l=1}, \ \bm{y}_{k,l}  \in \mathbb{R}^3$.
These measurements are generated by the corresponding measurement source on the shape surface $\bm{z}$:
\begin{equation}
\label{eq: y}
    \bm{y}_{k,l} = \bm{z}_{k,l} + \bm{v}_{k,l}, \quad \bm{v}_{k,l} \sim \mathcal{N}(\bm{0}, \bm{R}_{k,l}),
\end{equation}
where $\bm{v}_{k,l}$ is the measurement noise and $\bm{R}_{k,l}$ is the corresponding covariance matrix.
Given the fact that the association of measurement to source is not trivial, we follow the level-set \cite{levelset} implicit measurement formulation. 
\begin{equation}
\label{eq: levelset}
    \mathcal{L}_\phi(\bm{x_k},t) = \{ \bm{z}_{k,l} | \phi(\bm{x}_k, \bm{z}_{k,l})=t\} 
\end{equation}
describes the level-set for level $t \in \mathbb{R}$ \cite{levelset} . 
Using the extrusion model described in Eq. (\ref{eq: shapemodel}), we consider only the points at the boundary, leaving $t$ at $0$.
We combine Eq. (\ref{eq: y}) and Eq. (\ref{eq: levelset}) to get 
\begin{equation}
    \begin{split}
    0&=t - \phi(\bm{x}_k, \bm{y}_{k,l} - \bm{v}_{k,l}) \\
    &=0 - \phi(\bm{x}_k, \bm{y}_{k,l} - \bm{v}_{k,l}) \\
    &= h(\bm{x}_k, \bm{y}_{k,l}, \bm{v}_{k,l})
    \end{split}.
\end{equation}
Furthermore, we use the $x,z$ component of the points on the extrusion, as well as the $y$ components of the points in the caps. 
The pseudo-measurement 0 then has the representation as follows:
\begin{equation}
\label{eq:h}
    \begin{split}
    \bm{0}&=\bm{h}=[h_{e,x} \ h_{e,z}, \ h_{c,y}]^\top,\\
        h_{e,x}&=\left(T^{-1}[\bm{y}_{k,l}^\top \ ,1]^\top\right)_x - \bm{s}_x(\tau_{k,l}) - \bm{v}_{k,l,x}\\
        &=\mathrm{c}(\psi_k) y_{k,l,x} + \mathrm{s}(\psi_k)y_{k,l,y} - \mathrm{c}(\psi_k)x_{k,x}\\
        &\quad - \mathrm{s}(\psi_k)x_{k,y} - \bm{s}_x(\tau_{k,l}) - \bm{v}_{k,l,x}\\
        &=\mathrm{c}(\psi_k) y_{k,l,x} + \mathrm{s}(\psi_k)y_{k,l,y} - \mathrm{c}(\psi_k)x_{k,x}\\
        &\quad - \mathrm{s}(\psi_k)x_{k,y} -     \sum_{i = 1}^{n} B_{i,d}(\tau_{k,l})x_{c_x^i} - \bm{v}_{k,l,x},\\        
        h_{e,z}&=\left(T^{-1}[\bm{y}_{k,l}^\top \ ,1]^\top\right)_z - \bm{s}_z(\tau_{k,l}) - \bm{v}_{k,l,z}\\
        &=y_{k,l,z} -x_{k,z}- \bm{s}_z(\tau_{k,l}) - \bm{v}_{k,l,z}\\
        &=y_{k,l,z} -x_{k,z}- \sum_{i = 1}^{n} B_{i,d}(\tau_{k,l})x_{c_z^i} - \bm{v}_{k,l,z},\\
        h_{c,y}&=\left(T^{-1}[\bm{y}_{k,l}^\top \ ,1]^\top\right)_y - \left(\pm \frac{q}{2}\right) - \bm{v}_{k,l,y}\\
        &=-\mathrm{s}(\psi_k) y_{k,l,x} + \mathrm{c}(\psi_k)y_{k,l,y} + \mathrm{s}(\psi_k)x_{k,x}\\
        & \quad- \mathrm{c}(\psi_k)x_{k,y}
         - \left(\pm \frac{q}{2}\right) - \bm{v}_{k,l,y},\\
    \end{split}
\end{equation}
where $T$ is the transformation matrix from the global frame of reference to the body frame of reference of the object.
Eq. (\ref{eq: local measurement source}) is substituted into the equation.
For a detailed treatment of the pseudo-measurement, we refer the reader to \cite{levelset}.

For each measurement in the sensor data associated with the shape model, we must first assign the points on the extrusion and caps, and then find the $\tau$ for the points on the contour $\bm{s}$.
To check if a point belongs to the extrusion part, we compute a convex hull \cite{convexhull} of the measurements in the body reference frame. 
The vertices on the convex hull are used as points on the extrusion. 
For a more detailed contour, an alphashape \cite{alphashape} with $\alpha > 1$ could be used.
These points are further mapped to the estimated shape of the B-spline curve $\bm{s}$ by calculating their minimum distance to the curve. 
\begin{equation}
    \tau = \arg \min_\tau (\| \bm{z}^L_{xz} - \bm{s}(\tau) \|)
\end{equation}
The points on the caps are then found by filtering all points with $z^L_{y} > \lambda q/2$. 
In addition to the measurement models above, we add another constraint. 
The control points of the B-spline curve are allowed to move freely in space. 
For convenience, we let the B-spline curve interpolate the first and last control points. 
Therefore, we have the node vector for the $n$ control points as 
\begin{equation}
\begin{split}
    \bm{\tau} &= (\tau_i)_{i=1}^{n+d+1}\\ &= [\underbrace{0,\cdots,0}_{d+1},1,2,\cdots, \underbrace{n-d,\cdots,n-d}_{d+1}
\end{split}
].
\end{equation}
We further fix the first and the last control points on the same height, which indicates a closure in $z$ direction. 
These two control points though are allowed to move freely in $x$ direction.
In the case where the sensor data is mainly collected from the chassis in the upper part, the bottom of the object might not be estimated accurately, but could be represented by a line segment between these two control points.
We therefore have another pseudo-measurement 
\begin{equation}
        0 = x_{c^1_z}-x_{c^n_z} + v_{c_1, c_n},
\end{equation}
with the measurement noise $v_{c_1, c_n}$. 
\section{Evaluation}
\label{Chapter5: Evaluation}

We evaluate the method with both simulative data and real world open dataset. 
\subsection{Evaluation in Simulation}
\begin{figure}[!h]
    \centering
        \begin{tikzpicture}
        \node[anchor=south west,inner sep=0] (image) at (0,0) 
            {\includegraphics[width=\linewidth]{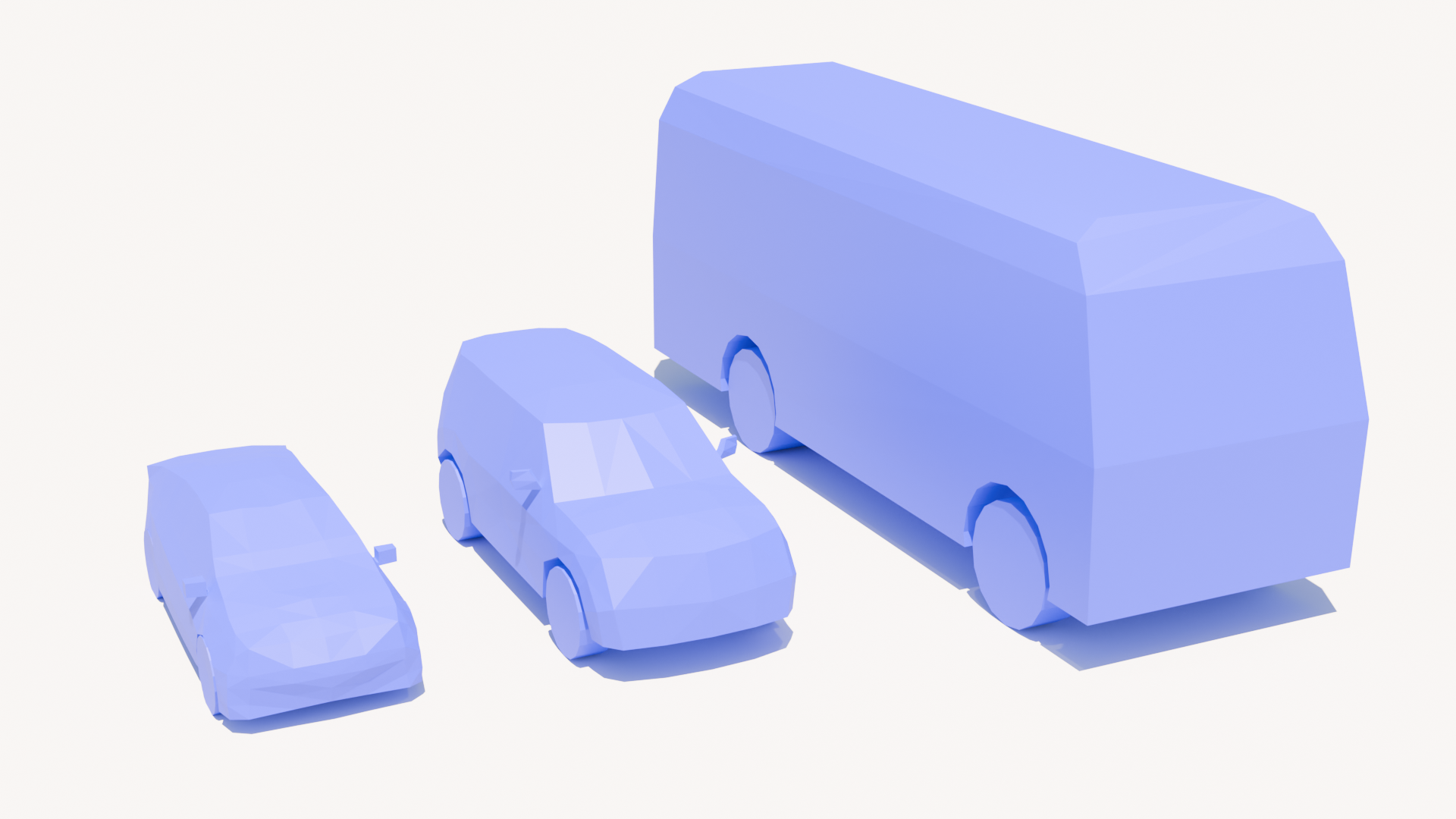}};
        \begin{scope}[x={(image.south east)},y={(image.north west)}]
            \draw[black, thick, ->] (0.15, 0.55) node[above=0pt] {Vehicle a} -- (0.18, 0.40);
           \draw[black, thick, ->] (0.35, 0.7) node[above=0pt] {Vehicle b} -- (0.38, 0.55);           \draw[black, thick, ->] (0.8, 0.85) node[right=0pt] {Vehicle c} -- (0.7, 0.8);              
        \end{scope}
    \end{tikzpicture}
    \caption{The vehicle models used in the simulation test.}
    \label{fig:vehicle models}
\end{figure}
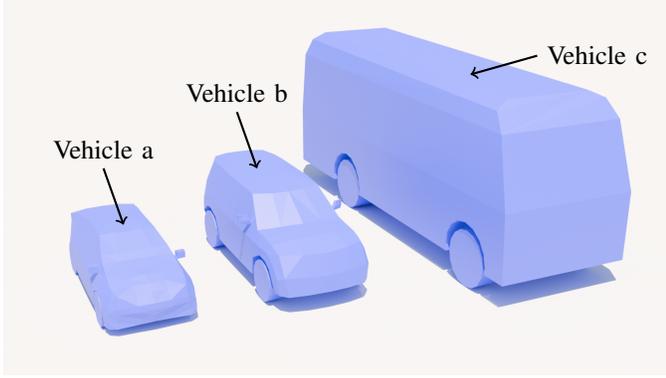

We use CARLA \cite{carla} to simulate the scenario shown in the Fig. \ref{fig:setup}.
The figure shows the position of the sensors, their field of view, the trajectory of the vehicle, and also the driving behavior along the trajectory.
A radar and a lidar are installed at each sensor position.
In the evaluation, we use a sampled point cloud from the ground truth mesh as a baseline measurement, and radar, lidar data to evaluate influences with perspectives and different data distributions from the sensor principle.
Both sensors operate at 10Hz.
The lidar is able to generate 1 million points per second. 
These points range from 5° to -60°, with the number of layers being 256.
The radar can generate half a million points per second. 
It has a vertical FOV of 60°.
Both sensors are allowed to get measurements up to 200 $m$.
Furthermore, we use different types of vehicles as shown in Fig. \ref{fig:vehicle models}.

\begin{figure}[h]
    \centering
    \includegraphics[trim={2cm 1.7cm 2cm 1.7cm},clip, width=1\linewidth]{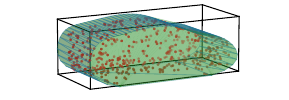}
    \caption{An example result of one frame. 
    The red dots are the sampled point cloud as measurements.}
    \label{fig:3d representation}
\end{figure}

Fig. \ref{fig:3d representation} depicts an example of the tracking result.
Here the red points are the measurements.
The green mesh is the estimation.
For the motion evaluation, we compare the state with the motion and orientation of the ground truth bounding box (shown as a black frame in the figure).
For the evaluation of the shape estimation we compare the estimated mesh with the ground truth mesh (not shown in the figure).

\begin{figure*}
    \centering
    \includegraphics[width=1\linewidth]{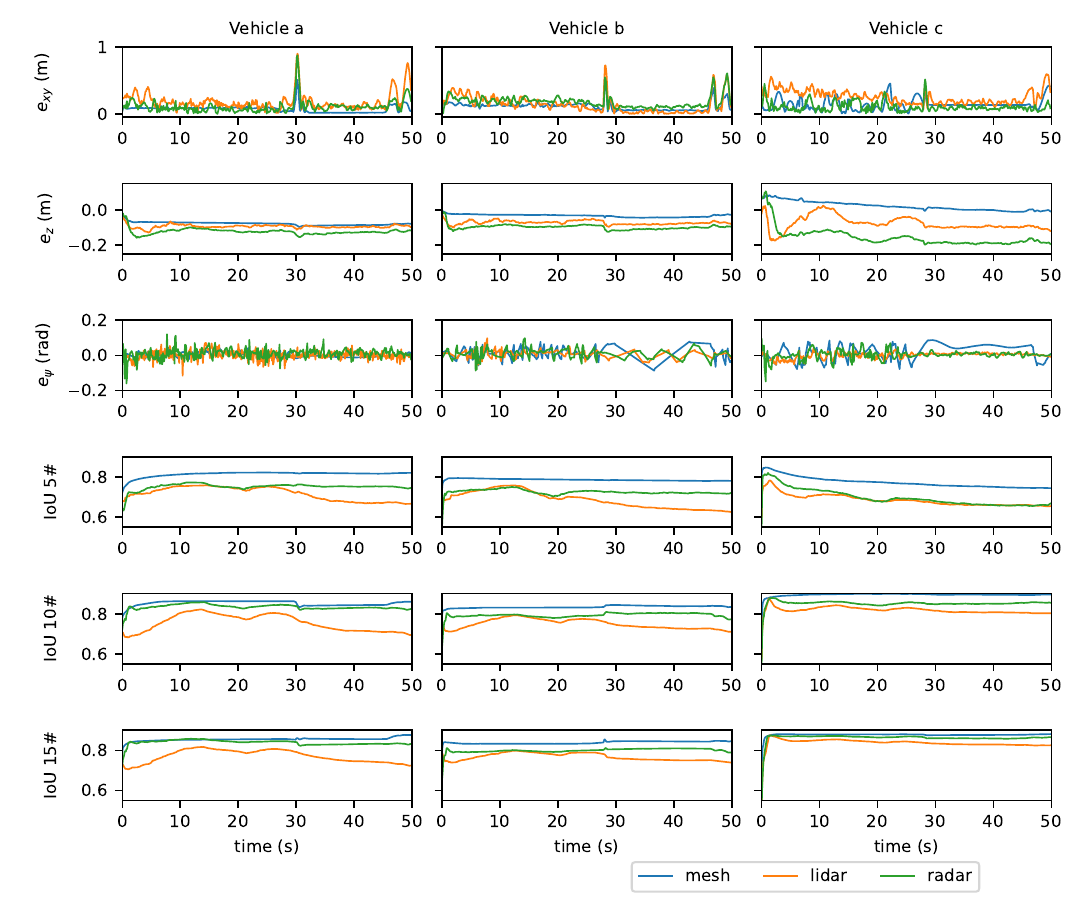}
    \caption{The result of state estimation for different vehicles.
    The results with different measurement data for all three models are shown. 
    The shape estimation evaluation is shown with different numbers of control points of the B-spline curve.}
    \label{fig:simulation test}
\end{figure*}

First, we evaluate the tracking result in terms of motion.
As shown in the first three rows of Fig. \ref{fig:simulation test}, the tracker is able to track all three vehicle models with high accuracy. 
The tests are performed with 10 control points of a 3-degree B-spline curve as the shape model.
The measurement noise $\sigma_m$ in $\bm{R}_{k,l}$ is set to $0.5$.
The process noise is $\sigma_x = \sigma_y=0.5\cdot8.8\cdot\Delta t^2, \sigma_{v_{xy}}=8.8\cdot\Delta t, \sigma_\psi=0.1\cdot\Delta t, \sigma_\omega=\Delta t, \sigma_z=0.1\cdot\Delta t, \sigma_{v_z}=0.01 \cdot\Delta t$, respectively. 
The process noise of the extent information is set to $0.1$.
The errors are calculated between the estimate and the center and yaw angle of the bounding box. 

We now go through the results for vehicle a to c and with the sampled point cloud, radar data and lidar data.
Vehicle a is the smallest model and thus get a compact measurement in the tracking process by the four sensors. 
The error in 2D ground motion plane is at most period about 0.2~$m$. 
The vertical error is about 0.1~$m$. 
The orientation can be kept close to 0 $rad$, with occasional peaks up to 0.1~$rad$. 
This happens e.g. before 10~$s$, when the vehicle makes the lane change. 
In the left turn period, we also observe that the orientation can be kept at correct values, as shown in the period 10~$s$ - 20~$s$. 
However, the dynamic behavior, such as braking, gives a large perturbation in the ground plane movement, as shown around 30~$s$. 
We can also see that the movement can be better tracked using the sampled point cloud, where the errors are only up to 0.5~$m$. 
In the cases using radar or lidar data, we could have a large error in the 2D displacement up to about 1~$m$. 
Furthermore, we observe that using radar data in CARLA the result is better than the lidar in the 2D motion, indicating the fact that the radar point cloud could still deliver good results when the vehicles move away from the sensors, (e.g. before 10~$s$), although it could generate less total number of measurements. 
lidars, on the other hand, may work well when the vehicle is close, but may not provide representative measurements at a distance due to the loss of density.
Another insight would be that the errors in $z$ when using lidar are smaller than when using radar, indicating a wider and more accurate distribution of points in the vertical direction. 

Vehicle b shows similar results to vehicle a, given a larger volume of the geometric model. 
However, vehicle c has some differences. 
For the 2D ground motion, although the waiting phase (30~$s$-45~$s$) is similarly characterized by small positional error, the estimation gives larger errors up to 0.5~$m$. 
More interestingly, tracking with lidar data gives larger errors than tracking with radar data. 
A difference lies in the perception of the start position, for the smaller models the vehicle could be seen as a whole, but for the bus model the measurements do not cover all parts of the vehicle. 
This leads to the initial bias. 
After a while, the radar points could give a sparse but comprehensive description, but the lidar points become less representative, which leads to a larger error in the run. 
The height error is larger in the radar and lidar cases, going up to 0.2~$m$. 
This could be due to the fact that the underside of the vehicle is less observed. 

In addition to the motion evaluation, we evaluate the shape estimation with the Intersection Over Union (IoU) on the side view, from 0 to 1, the higher the better. 
The ground truth is obtained from the mesh.
The B-spline curve is closed by connecting the first and last control points to form a polygon. 
The shape estimation all starts with an arc shaped initial model, for vehicle a and b the radius is 2~$m$, for the bus the radius is 4~$m$.
The results are shown in the last three rows of Fig. \ref{fig:simulation test}.
The tests are performed with the same tracking parameters except for the number of control points used in the B-spline curve.
5, 10 and 15 control points are tested. 
For all three vehicles, we can clearly see that the curves with 10 or 15 control points generally perform better than shape models with only 5 control points. 
As expected, the sampled point cloud shape estimation gives the best result. 
The radar point cloud shape estimation again outperforms the lidar point cloud shape estimation, as shown before, due to the data distribution.
After the vehicle stops at the traffic light, the shape estimation of vehicle a experiences a drop in accuracy, mainly due to the slow response in motion to the dynamic change.
However, after the vehicle b stops, the IoU has a slight jump in the positive direction, resulting from a bending behavior of the estimation over the hood. 
The change in motion causes some control points to be "squeezed" together, resulting in a counterintuitive but beneficial result.
The shape estimation on the side view of the bus is very accurate due to the relatively simple geometry of the bus. 
The IoU can sometimes reach 0.9. 
The tests with the sampled point cloud show the best result, since the wheels and other details could be captured.

\subsection{Evaluation in real world dataset}
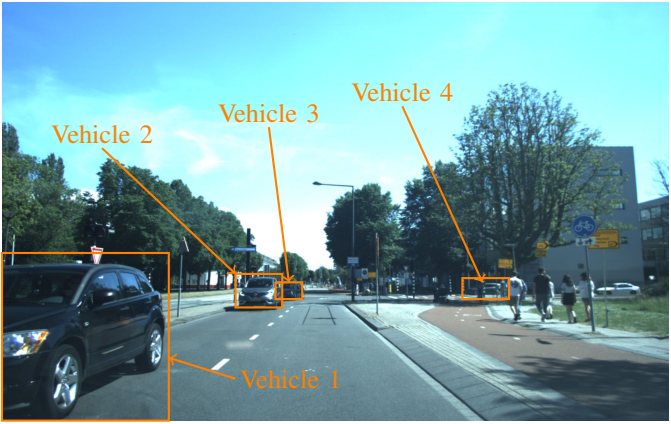
\begin{figure}[!h]
    \centering
        \begin{tikzpicture}
        \node[anchor=south west,inner sep=0] (image) at (0,0) 
            {\includegraphics[width=\linewidth]{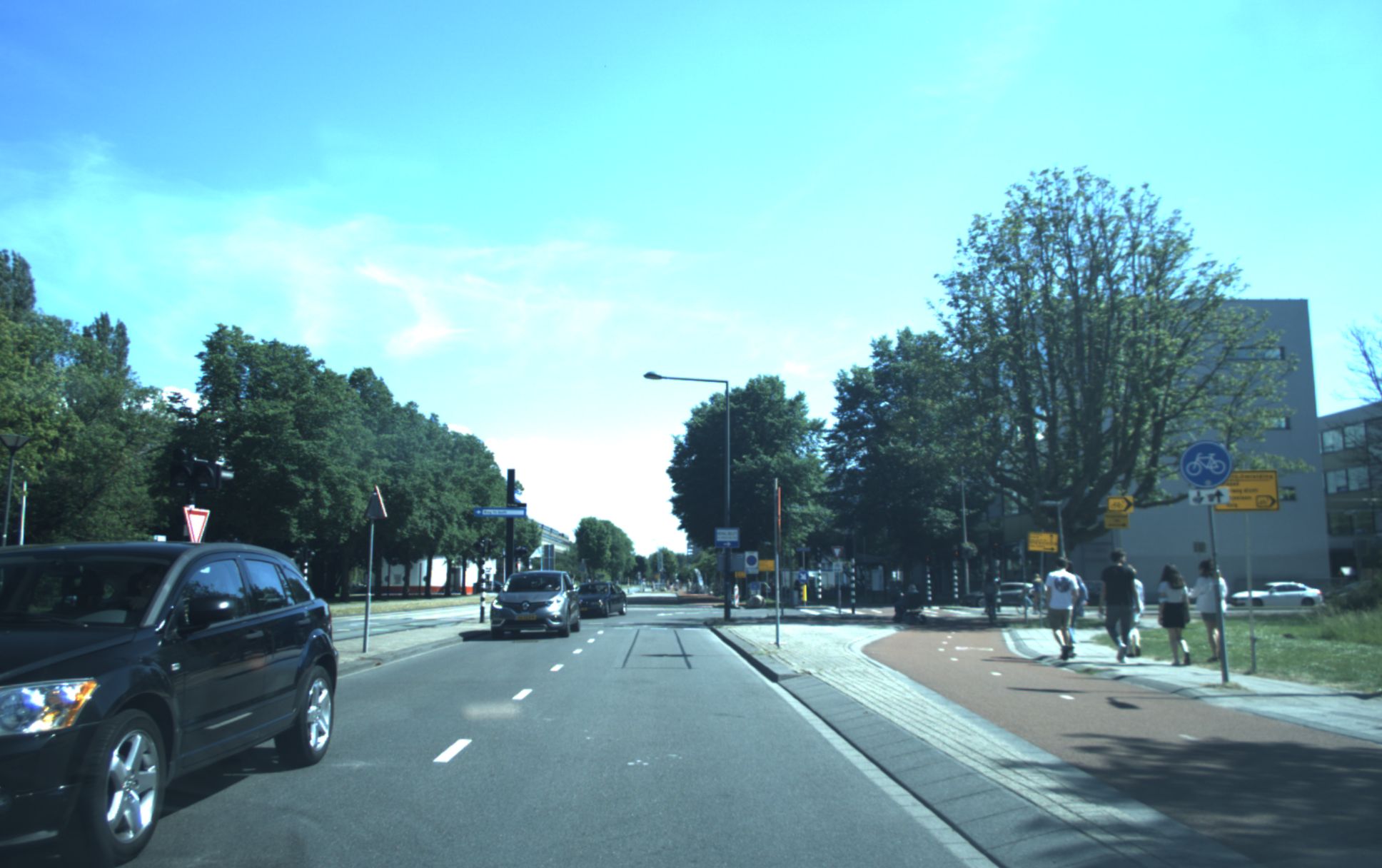}};
        \begin{scope}[x={(image.south east)},y={(image.north west)}]
            \draw[orange, thick, ->] (0.35, 0.1) node[right=-2pt] {Vehicle 1} -- (0.25, 0.15);
            \draw[orange, thick] (0,0) rectangle (0.25,0.4);
            \draw[orange, thick, ->] (0.15, 0.65) node[above=-2pt] {Vehicle 2} -- (0.35, 0.35);
            \draw[orange, thick] (0.35,0.27) rectangle (0.42,0.35);
            \draw[orange, thick, ->] (0.4, 0.7) node[above=-2pt] {Vehicle 3} -- (0.43, 0.33);
            \draw[orange, thick] (0.41,0.29) rectangle (0.45,0.33);     
            \draw[orange, thick, ->] (0.6, 0.75) node[above=-2pt] {Vehicle 4} -- (0.72, 0.33);
            \draw[orange, thick] (0.69,0.29) rectangle (0.76,0.34);     
        \end{scope}
    \end{tikzpicture}
    \caption{The test scenario in the real word dataset \cite{vod}}
    \label{fig:vod vehicles}
\end{figure}
We further use a real world dataset, the view-of-delft dataset \cite{vod} for the evaluation. 
We choose to use this dataset for both 3D radar and lidar evaluation.
The four vehicles are shown in Fig. \ref{fig:vod vehicles}. 
Vehicles 1 to 3 are seen as incoming objects with straight line motion. 
Vehicle 4 is seen as an object making a left turn. 
The geometry of these vehicles is all similar to vehicle a in the simulated test, so the same tracking parameters are used.
We use 10 control points for shape estimation.

\begin{figure*}[b]
    \centering
    \includegraphics[width=1\linewidth]{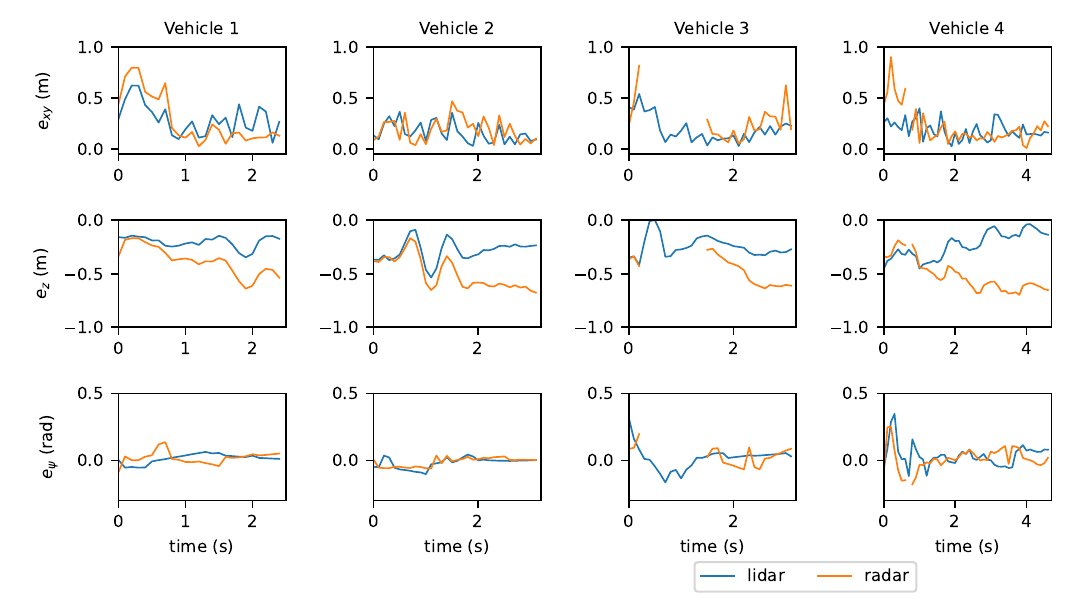}
    \caption{The result of state estimation of different vehicles in the real dataset sequences.
    The results with different measurement data for all 4 vehicles are shown. }
    \label{fig:vod test}
\end{figure*}
The result is summarized in Fig. \ref{fig:vod test}. 
The positional error on the driving ground can be limited to less than 0.5 $m$.
The orientation could also be kept accurate, both in a straight ahead drive and a left turn.
The estimation of $z$ is less accurate, especially for radar tests.
We think this is due to the mounting position of the radar. 
It is at the level of the air intake, which leads to the problem of lower estimates in this scene.
For the vehicle 3 and 4, the radar plot is not consistent, this is because that the measurements of the sensor for these cars are less than 3 points. 
The tracker has skipped these frames.
It can be seen that the tracker could quickly recover the position of the car (see vehicle 3 at 1.5~$s$).
lidar tests show a good overall performance. 
The shape estimation cannot be evaluated directly because the ground truth only contains 3D bounding box information.

\section{Conclusion}
\label{Chapter6: Conclusion}
We propose a novel approach for tracking a 3D extended object, using the extrusion of the side view profile, which is represented by a B-spline curve.
We provide a through description of the shape model and derive the measurement model for the measurement source on the caps and on the extrusion.
To evaluate the method, we used extensive tests based on CARLA simulation data. 
The tests consist of sampled point cloud from ground truth mesh, simulated lidar and radar data. 
We also use sequences in open real-world datasets to evaluate the methods. 
Both tests show promising results.
The methods can be applied to autonomous driving scenarios and roadside intelligent transportation systems. 

Further development lies in using this method for distributed tracking tasks and understanding its behavior when fusing posterior estimates.
One limitation of this method that was observed in the tests was that the ambiguity of shape change and motion could lead to symmetric offsets in both the position and shape estimates, which normally add up to 0. 
These offsets are due to the flexibility in the placement of the control points of the B-spline curve. 
To limit this offset, internal constraints in the form of virtual forces could be designed between the control points and the position estimation.

\section*{Appendix}
From Eq. (\ref{eq:h}), we have 
\begin{equation}
    \frac{\partial h}{\partial \bm{x}}
= \left[ \frac{\partial h_{e,x}}{\partial \bm{x}} \ \frac{\partial h_{e,z}}{\partial \bm{x}} \ \frac{\partial h_{c,y}}{\partial \bm{x}}\right]^\top 
\end{equation}
\begin{equation}
    \frac{\partial h_{e,x}}{\partial \bm{x}}
= \left[ \frac{\partial h_{e,x}}{\partial x_x} \ \frac{\partial h_{e,x}}{\partial x_y} \ \frac{\partial h_{e,x}}{\partial \psi} \ \frac{\partial h_{e,x}}{\partial x_{c_x^i}} \right]
\end{equation}
\begin{equation}
    \frac{\partial h_{e,x}}{\partial x_x}
= - \mathrm{c}(\psi_k), \ \frac{\partial h_{e,x}}{\partial x_y}=- \mathrm{s}(\psi_k)  
\end{equation}
\begin{equation}
\begin{split}
    \frac{\partial h_{e,x}}{\partial \psi} =&-\mathrm{s}(\psi_k) y_{k,l,x} + \mathrm{c}(\psi_k)y_{k,l,y} \\
    &+ \mathrm{s}(\psi_k)x_{k,x} - \mathrm{c}(\psi_k)x_{k,y} 
\end{split}
\end{equation}
\begin{equation}
    \frac{\partial h_{e,x}}{\partial x_{c_x^i}}=B_{i,d}(\tau)
\end{equation}
\begin{equation}
    \frac{\partial h_{e,z}}{\partial \bm{x}}
= \left[ \frac{\partial h_{e,z}}{\partial x_z} \ \frac{\partial h_{e,z}}{\partial x_{c_z^i}} \right]^\top 
\end{equation}
\begin{equation}
    \frac{\partial h_{e,z}}{\partial x_z}=1, \ \frac{\partial h_{e,z}}{\partial x_{c_z^i}}=B_{i,d}(\tau)
\end{equation}
\begin{equation}
    \frac{\partial h_{c,y}}{\partial \bm{x}}
= \left[ \frac{\partial h_{c,y}}{\partial x_x} \ \frac{\partial h_{c,y}}{\partial x_y} \ \frac{\partial h_{c,y}}{\partial \psi}  \right]^\top 
\end{equation}
\begin{equation}
    \frac{\partial h_{c,y}}{\partial x_x}
= \mathrm{s}(\psi_k), \ \frac{\partial h_{c,y}}{\partial x_y}=- \mathrm{c}(\psi_k)  
\end{equation}
\begin{equation}
\begin{split}
    \frac{\partial h_{c,y}}{\partial \psi} =&\mathrm{c}(\psi_k) y_{k,l,x} - \mathrm{s}(\psi_k)y_{k,l,y} \\
    &- \mathrm{c}(\psi_k)x_{k,x} + \mathrm{s}(\psi_k)x_{k,y} 
\end{split}
\end{equation}

\bibliographystyle{IEEEtran}
\bibliography{refs}

\end{document}